\documentclass[sigconf]{acmart}

\usepackage[utf8]{inputenc}
\usepackage{listings}
\usepackage{multirow}
\usepackage[linesnumbered,ruled,vlined]{algorithm2e}
\usepackage{dirtree}

\usepackage{longtable}
\usepackage{lipsum}
\usepackage{enumitem}
\usepackage{balance}

\AtBeginDocument{%
  }

\copyrightyear{2025}
\acmYear{2025}
\setcopyright{acmlicensed}
\acmConference[MM '25] {Proceedings of the 33rd ACM International Conference on Multimedia}{October 27--31, 2025}{Dublin, Ireland.}
\acmBooktitle{Proceedings of the 33rd ACM International Conference on Multimedia (MM '25), October 27--31, 2025, Dublin, Ireland}
\acmISBN{979-8-4007-2035-2/2025/10}
\acmDOI{10.1145/3746027.3755418}

\acmSubmissionID{3665}

\begin{document}

\title{Tree-of-Reasoning: Towards Complex Medical Diagnosis via Multi-Agent Reasoning with Evidence Tree}

\author{Qi Peng}
\authornote{Equal contribution.}
\affiliation{%
  \institution{South China University of Technology}
  \city{Guangzhou}
  \country{China}}
\affiliation{%
  \institution{Key Laboratory of Big Data and Intelligent Robot (SCUT), Ministry of Education}
  \city{Guangzhou}
  \country{China}}
\email{sepengqi@mail.scut.edu.cn}

\author{Jialin Cui}
\authornotemark[1] 
\affiliation{%
  \institution{South China University of Technology}
  \city{Guangzhou}
  \country{China}}
\email{202130140117@mail.scut.edu.cn}

\author{Jiayuan Xie}
\affiliation{%
  \institution{Hong Kong Polytechnic University}
  \city{Hong Kong}
  \country{China}}
\email{jiayuan.xie@polyu.edu.hk}

\author{Yi Cai}
\authornote{Corresponding author.}
\affiliation{%
  \institution{South China University of Technology}
  \city{Guangzhou}
  \country{China}}
\affiliation{%
  \institution{Key Laboratory of Big Data and Intelligent Robot (SCUT), Ministry of Education}
  \city{Guangzhou}
  \country{China}}
\email{ycai@scut.edu.cn}

\author{Qing Li}
\affiliation{%
  \institution{Hong Kong Polytechnic University}
  \city{Hong Kong}
  \country{China}}
\email{qing-prof.li@polyu.edu.hk}

\renewcommand{\shortauthors}{Qi Peng, Jialin Cui, Jiayuan Xie, Yi Cai, \& Qing Li}

\begin{abstract}
Large language models (LLMs) have shown great potential in the medical domain. However, existing models still fall short when faced with complex medical diagnosis task in the real world. This is mainly because they lack sufficient reasoning depth, which leads to information loss or logical jumps when processing a large amount of specialized medical data, leading to diagnostic errors. To address these challenges, we propose Tree-of-Reasoning (ToR), a novel multi-agent framework designed to handle complex scenarios. Specifically, ToR introduces a tree structure that can clearly record the reasoning path of LLMs and the corresponding clinical evidence. At the same time, we propose a cross-validation mechanism to ensure the consistency of multi-agent decision-making, thereby improving the clinical reasoning ability of multi-agents in complex medical scenarios. Experimental results on real-world medical data show that our framework can achieve better performance than existing baseline methods\footnote{The source code can be found in https://github.com/tsukiiiiiiiii/TOR.git.}.

\end{abstract}




\maketitle

\section{Introduction}
In recent years, large language models (LLMs) have demonstrated remarkable capabilities across a wide range of tasks \cite{yu2024towards,li2024econagent,li2024empowering,yuan2024few}.
Inspired by their successful applications in fields such as natural language processing, LLMs have begun attracting substantial attention in the medical domain \cite{yang2023large}, where medical diagnosis based on LLMs has emerged as particularly promising \cite{zhou2024large,goh2024large}. By incorporating extensive medical knowledge and sophisticated reasoning capabilities, LLMs can automatically diagnose numerous diseases, which not only enhance the efficiency of clinical decision-making but also reduce the burden on doctors. 

Many medical diagnostic methods based on LLM have been proposed, such as MINIM \cite{wang2025self}, CHIEF \cite{wang2024pathology}, and HealthGPT \cite{lin2025healthgpt}.
These medical LLMs are primarily specialized models, designed to analyze and provide diagnoses for a single type of medical examination data within specific domains.
However, real-world medical diagnosis is often more challenging. A single type of data is often insufficient to capture the full spectrum of a disease’s clinical manifestations. For example, certain diseases such as tumors typically require a combination of radiology, laboratory tests, and pathological data for accurate diagnosis. Therefore, doctors usually need to integrate multiple types of medical examination data to make a comprehensive clinical diagnosis.
To address these complex diagnostic tasks, researchers have introduced medical-oriented multi-agent methods \cite{tang2024medagents,kim2024mdagents,peng2024integration}. The multi-agent frameworks can simulate the collaborative process of multiple experts, where each agent can focus on processing specific data. 


Although multi-agent methods enable better integration of multi-source information to address complex situations, they still suffer from the following limitations:
(1) \textbf{Insufficient interpretability in medical agents:} Medical diagnoses in the real-world usually require a clear reasoning process and sufficient evidence chain to support them, however, existing multi-agent approaches only provide diagnosis results with some simple explanations. This ``black box'' decision-making method is unacceptable in the medical domain, where rigor and accountability are essential. The lack of interpretability seriously restricts its application in actual medical scenarios.
(2) \textbf{Multi-agent reasoning conflict}: Different agents may have different reasoning perspectives or strategies, which can lead to conflicting evidence or contradictory conclusions during diagnosis process. 
Such reasoning conflict increases the uncertainty of the overall system \cite{weng2020uncertainty,tang2021collaborative}, thereby raising the risk of misdiagnosis and potentially endangering patient health.

\begin{figure}[t]
  \centering
  \includegraphics[width=\linewidth]{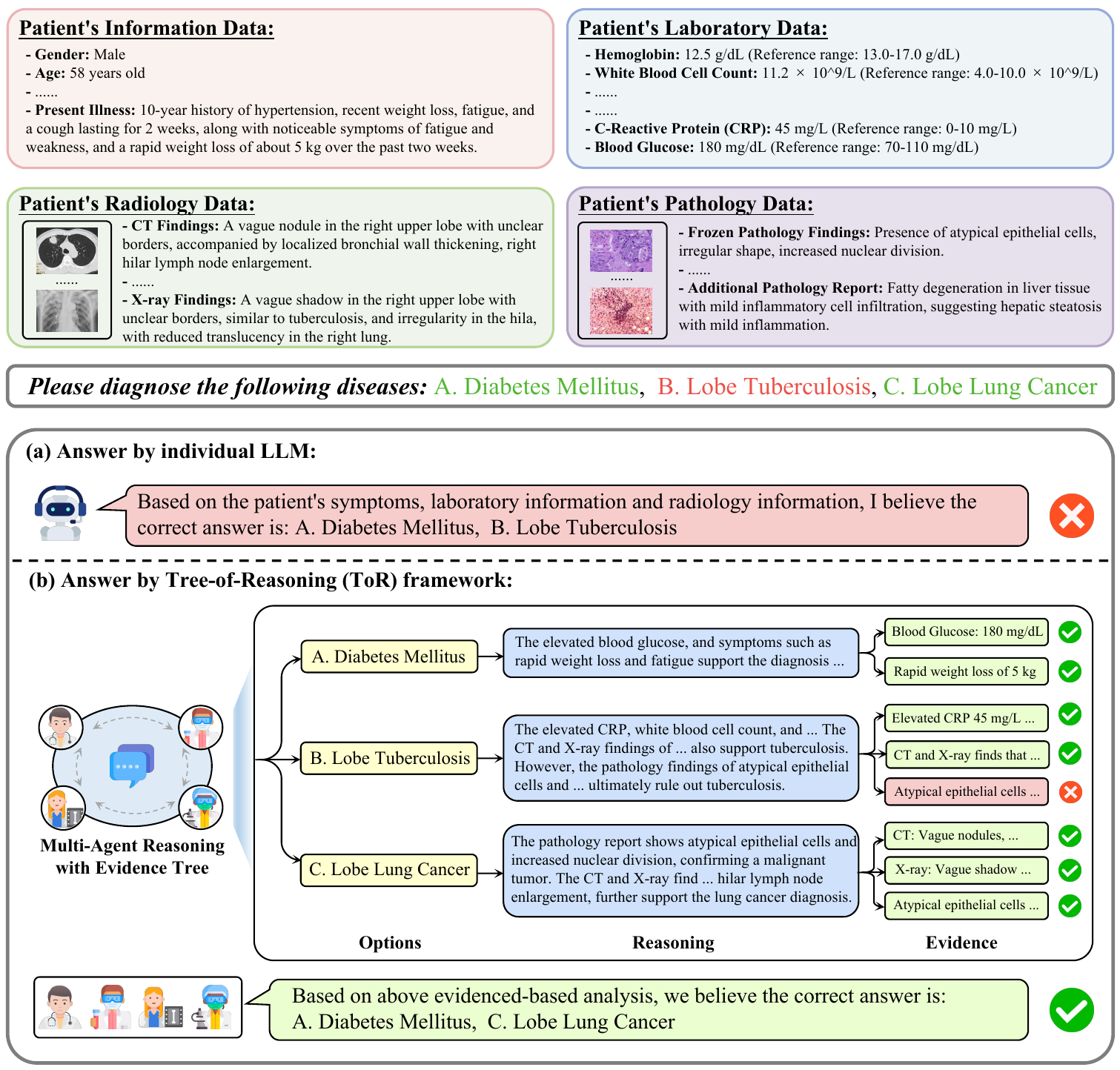}
  \caption{Overview of Tree-of-Reasoning (ToR) multi-agent framework, which diagnoses complex medical scenarios through reasoning paths with evidence tree.}
  \label{fig:intro}
\end{figure}

To overcome the above limitations, we propose a framework called \textbf{Tree-of-Reasoning (ToR)} for multi-agent collaborative diagnosis, which is shown in Figure~\ref{fig:intro}. 
To address the first limitation, we follow the ``evidence-based medicine'' theory 
\cite{sackett1997evidence}, 
the core of which is to guide diagnosis based on the most relevant clinical evidence. When the agent simulates the process, it needs to show specific reasoning steps and evidence chains, thereby transforming the originally “black-box” decision-making process into a transparent one.
Specifically, to capture the relationship between diagnosis, reasoning and evidence, we propose a tree-structured reasoning path. This tree structure can clearly display the reasoning process and make each decision step supported by clinical evidence, thereby enhancing the interpretability of reasoning and reducing diagnostic bias caused by logical jumps or insufficient evidence.
To address the second limitation, we refer to relevant researches on multi-agent debate \cite{liang2023encouraging,chan2023chateval}, which allows agents with conflicting conclusions to discuss with each other, promoting the identification, evaluation and resolution of conflicts. Following this, we propose a multi-agent cross-verification mechanism, in which different agents can review the diagnostic reasoning paths and evidence chains of other agents from their own observation perspectives. This mutual review process helps to discover potential conflicts, errors or information omissions, thereby promoting consensus among agents and improving the accuracy of overall diagnosis.

In our framework, each agent is acted as a specialist to process corresponding medical data from its own perspective. Specifically, we design four types of agents based on the classification of real-world medical data, i.e., outpatient doctor, laboratory doctor, radiology doctor, and pathology doctor. In detail, the outpatient doctor outputs diagnostic reasoning based on the patient's symptom history; the laboratory doctor outputs diagnostic reasoning based on laboratory test indicators; the radiology doctor outputs diagnostic reasoning based on radiology findings; and the pathology doctor outputs diagnostic reasoning based on pathology findings. The output of each agent includes the detailed reasoning process and corresponding evidence in a tree structure to improve interpretability. Subsequently, the cross-verification mechanism allows each agent to review and correct each other’s diagnosis, reducing multi-agent reasoning conflict and making a consistent diagnosis.


Considering that most of the existing medical datasets are single-source medical data, To evaluate different methods under complex medical diagnosis scenario, we collect real patient data from a real-world hospital, which included patient information (such as patient gender, age, and current medical history), laboratory test indicators (such as blood routine indicators and carcinoembryonic protein indicators), radiology reports (such as MRI, CT, and X-ray findings), and pathology reports (such as tumor grade and stage). The experimental results demonstrate that ToR framework can achieve comparable or superior performance to existing baseline methods. Further analysis indicates that ToR is more practical in clinical decision support.
We summarize our contributions as follows: 
\begin{itemize}[itemsep=2pt,topsep=0pt,parsep=0pt,leftmargin=0.5cm]

\item We propose Tree-of-Reasoning (ToR), the first multi-agent medical diagnosis framework that combines medical reasoning paths with clinical evidence to improve the diagnostic performance of agents in complex medical scenarios. 

\item We conduct comprehensive experiments that demonstrate our framework can surpass other baseline methods on the proposed datasets. 

\item Further analysis verifies that real doctors prefer to use our system compared to the baseline methods. This is of great significance for the development of AI-assisted medical care.

\end{itemize}

\section{Related Work}

\subsection{Large Language Models in Medical Domain}
Large Language Models (LLMs) have been applied across various domains, such as software development \cite{hou2024large,fan2023large}, finance \cite{li2023large,wu2023bloomberggpt}, gaming \cite{gallotta2024large,hu2024survey}, etc. Recently, LLMs have demonstrated significant potential in medical domain, including medical question answering \cite{singhal2025toward,lucas2024reasoning}, report generation \cite{liu2024bootstrapping,zhong2023chatradio}, clinical diagnosis \cite{wang2023chatcad,li2023chatdoctor,wen2024mindmap,peng2025cke}. Within medical LLMs, there are two main approaches: (1) training with domain-specific data for specialized models \cite{wang2023chatcad}, and (2) leveraging large general-purpose models with prompt engineering and retrieval-augmented generation (RAG) \cite{wen2024mindmap}.
Early research mainly focus on pre-training and fine-tuning strategies using medical knowledge. However, with the rise of large-scale general-purpose LLMs, training-free approaches have become feasible, which effectively tap into the models' inherent and external medical knowledge. For instance, GPT-4 \cite{achiam2023gpt}, with refined prompt tuning, outperforms specialized fine-tuned models like Med-PaLM \cite{nori2023capabilities}.
While large-scale general-purpose LLMs have shown promising results in general medical tasks, they fall short in complex medical diagnosis scenarios, which demand collaborative inference across multiple medical tasks \cite{peng2024integration,kim2024mdagents,tang2024medagents}. A single LLM may not fully capture the interdisciplinary dynamics inherent in complex medical diagnoses, which often need a multidisciplinary team (MDT). Therefore, we propose the integration of multiple specialized LLMs to collaborate effectively, thereby enhancing the accuracy and reliability of solving complex medical challenges.

\subsection{Multi-Agent Collaboration}
Research has shown that collaboration between multiple LLM agents can achieve better results than individual LLM in some complex scenarios, such as role-playing \cite{ye2025multi}, code generation \cite{hong2023metagpt,qian2024chatdev}, sociological simulations \cite{park2023generative}, and so on. Recently, multi-agent systems have been applied in medical domain \cite{peng2024integration,kim2024mdagents,tang2024medagents}. For instance, Tang et al.\cite{tang2024medagents} propose multidisciplinary collaboration (MC) framework, which uses LLM-based agents in a role-playing setting to participate in collaborative multi-round discussions. Kim et al.\cite{kim2024mdagents} introduce medical decision-making agents (MDAgents) framework, which enables dynamic collaboration among AI agents based on the complexity of medical tasks.
These works design different doctor-role agents, who engage in multiple rounds of discussion and make decisions to reach a final diagnosis. However, during these multi-round discussions, topic drift may occur, leading to the loss of key information \cite{becker2025stay,hong2023metagpt}. Furthermore, these approaches typically output the final diagnosis directly, lacking interpretability in the diagnostic process, which limits their utility in real-world medical decision-making.
To address these issues, inspired by the theory of ``evidence-based medicine'' \cite{sackett1997evidence}, we propose the multi-agent reasoning framework with evidence tree. This framework records the reasoning paths and evidence chains during the collaborative diagnostic process, improving both the accuracy and interpretability of multi-agent diagnosis.

\begin{figure*}[h]
  \centering
  \includegraphics[width=0.9\linewidth]{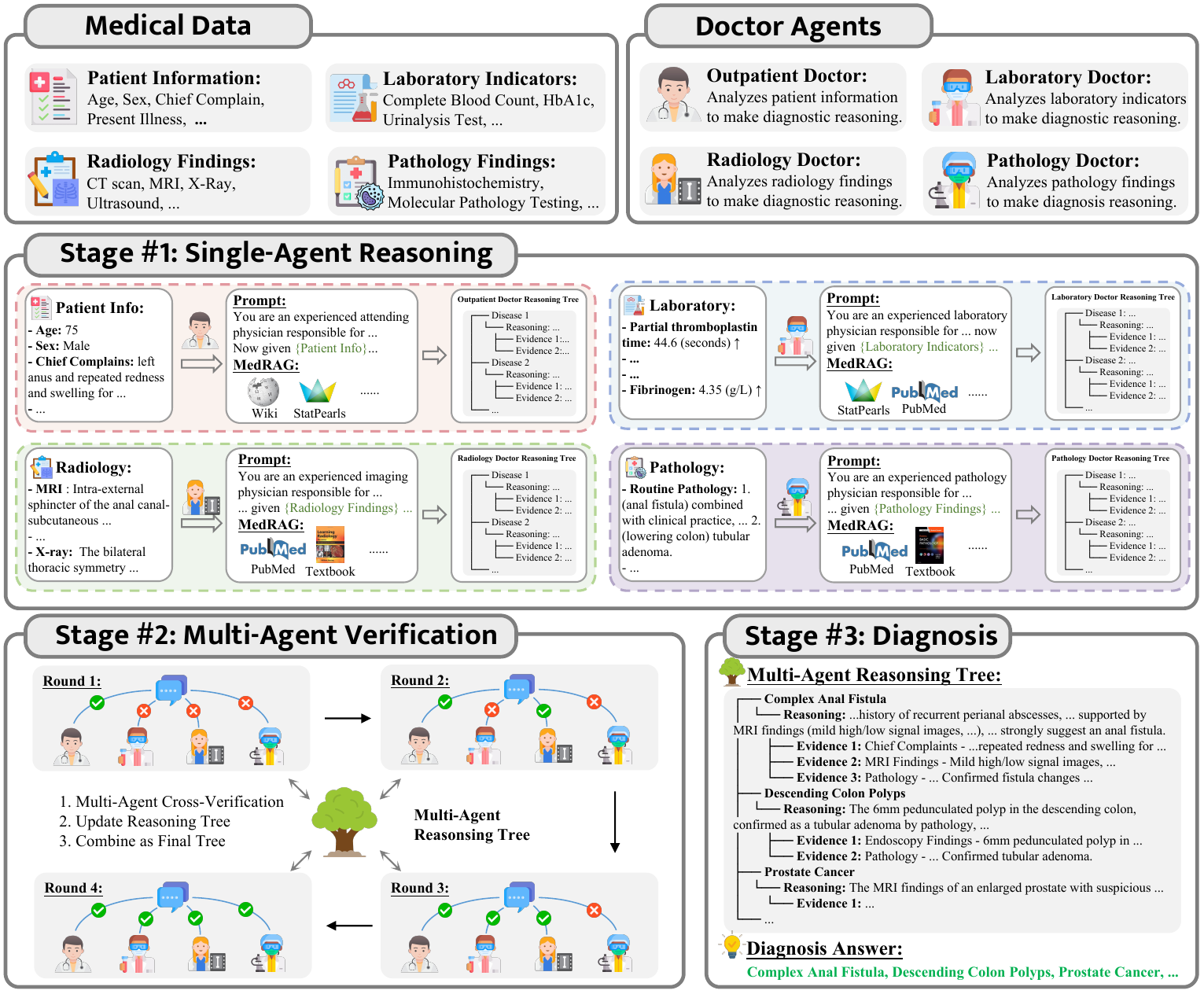}
      \caption{The overview of Tree-of-Reasoning framework. Initially, different doctor agents receive corresponding medical data and make diagnoses, outputting diagnostic paths with evidences. Next, the multi-agent verification mechanism corrects and updates the diagnostic paths of different agents. Finally, the diagnostic paths of all agents are summarized for the final diagnosis.}
  \label{fig:model}
\end{figure*}

\section{Methodology}
To handle diverse types of medical data and enable collaborative decision-making in complex medical diagnosis tasks, we propose an evidence-based multi-agent reasoning framework, named Tree-of-Reasoning (ToR). Unlike previous multi-agent systems, we construct a structured reasoning evidence tree following the theory of ``evidence-based medicine''. Specifically, each agent outputs its diagnosis, the reasoning path, and supporting evidence, which are organized as the first-level, second-level, and leaf nodes of the evidence tree, respectively. In addition, we design a multi-agent cross-verification mechanism. Through iterative discussions, agents reconcile conflicting diagnostic opinions to reach a consensus and jointly construct a final reasoning tree. The overview of ToR is showed as Figure~\ref{fig:model}.
We explain our framework in the following three parts:

\begin{itemize}[leftmargin=0.32cm]

\item \textbf{Specialization of Agents:} In this part, we define different doctor agents and introduce their responsibilities, as well as the medical RAG tool they use.

\item \textbf{Evidence-Based Reasoning Tree:} In this part, we explain the motivation for constructing the evidence-based reasoning tree, its detailed structure, and a specific example.

\item \textbf{Workflow:} In this part, we describe how the ToR framework executes diagnostic tasks and detailing the cross-verification mechanism leading to the final diagnosis.

\end{itemize}

\subsection{Specialization of Agents}
In real-world medical scenarios, patients often suffer from multiple diseases simultaneously. A comprehensive diagnosis typically requires the integration of various medical examination data and the expertise of different specialists. Inspired by this, in our framework, each agent is assigned a specific doctor role, focusing on particular tasks and contributing to a collaborative group diagnosis.

Considering that medical examination data can be divided into four main categories, including patient information, laboratory test data, radiology findings, and pathology findings, we design four corresponding doctor agents in our framework:

\noindent \textbf{Outpatient Doctor:}  This agent analyzes the patient's clinical symptoms, physical examination results, and past medical history, initially diagnosing common illnesses, prevalent diseases, and so on.
The reasoning path is usually based on clinical experience and typical manifestations of diseases.

\noindent \textbf{Laboratory Doctor:} This agent analyzes the patient's laboratory test results of blood, urine, body fluids, and secretions, diagnosing metabolic diseases, endocrine diseases, hematological diseases, and so on.
Their reasoning path relies on the abnormal changes in various biochemical indicators, immunological indicators, microbiological indicators.

\noindent \textbf{Radiology Doctor:} This agent analyzes the patient's medical imaging findings from X-rays, CT scans, MRIs, and ultrasounds, diagnosing oncological diseases, organ diseases, and other radiology diseases.
The reasoning path is based on radiological features, the morphology, size, and location of lesions, as well as their relationship with other tissues.

\noindent \textbf{Pathology Doctor:} This agent analyzes the patient's biopsy tissues, surgically removed specimens, and cytological smears, making the final diagnosis of tissue type and grade of tumors.
The reasoning path is based on morphological changes in cells and tissues combined with immunohistochemistry.

To enhance the reliability of the agent's diagnosis, every agent can refer to specialized medical resources in their domain through MedRAG \cite{xiong2024benchmarking} during data analysis. These resources include encyclopedic platforms such as Wikipedia, medical information databases like StatPearls, and medical textbooks. 
Each agent first analyzes the corresponding medical examination data to generate an initial diagnostic reasoning path. Then, through multiple rounds of discussion, the agents analyze different diagnostic conclusions from their respective professional perspectives. These agents engage in a collaborative decision-making process, contributing their expertise to reach a consensus and ultimately providing a final diagnosis for complex medical reasoning scenarios.
The specific prompts constructed by each doctor agent are detailed in the appendix of the supplementary material.

\subsection{Evidence-Based Reasoning Tree}


According to the theory of evidence-based medicine, every diagnosis should be supported by the most relevant clinical evidence \cite{sackett1997evidence}. When an agent simulates this process, it must explicitly present the reasoning path and corresponding clinical evidence, thereby transforming the black-box decision-making process into a transparent one and enhancing interpretability. Furthermore, prior studies have shown that large language models exhibit improved reasoning capabilities when the decision process is decomposed into step-by-step inference \cite{wu2025effectively,wei2022chain}. Based on this, we reasonably hypothesize that explicitly outputting the clinical reasoning path and recording the supporting clinical evidence will enhance the diagnostic capabilities of medical agents.

\begin{figure}[h]
  \centering
  \includegraphics[width=\linewidth]{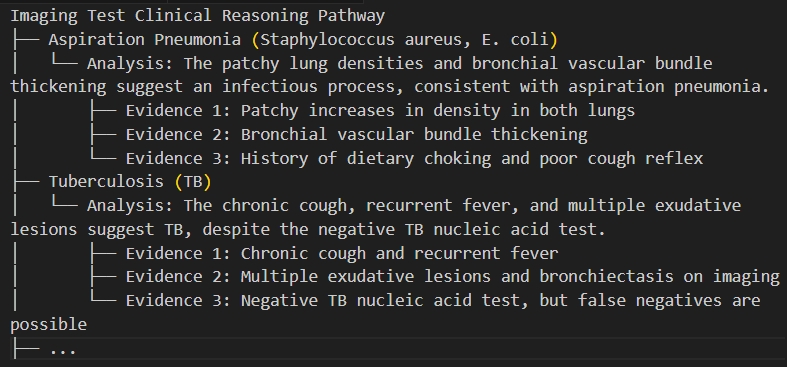}
  \caption{An example of the evidence-based tree structure.}
  \label{fig:tree_structure}
\end{figure}

To integrate the clinical reasoning paths of multiple agents, we have unified the format of different agents' paths, named ``evidence-based tree'', which aims to document the reasoning process corresponding to each diagnostic result, as well as the supporting clinical evidence for that reasoning.
Specifically, the evidence-based tree follows a hierarchical structure represented in a format similar to a markdown document. The first level consists of the diagnostic results, the second level contains the reasoning process for each diagnosis, and the third level provides the clinical evidence supporting the reasoning.
As shown in Figure~\ref{fig:tree_structure}, an example of the evidence-based tree structure is provided, where the first level shows diagnostic results such as Disease 1 and Disease 2, the second level presents the reasoning process for each diagnosis by the agent, and the third level lists the corresponding clinical evidence for each reasoning process, including patient symptoms, laboratory findings, imaging results, etc.

\subsection{Workflow}

In the workflow, each agent plays a specific role and performs tasks in a sequential manner as predefined. Specifically, after receiving different types of medical data, each agent carries out a distinct medical task and output reasoning path in a tree structure. The outpatient doctor analyzes the patient information, generating a diagnostic result and reasoning path from the outpatient doctor's perspective. The laboratory doctor examines the patient's laboratory indicators, producing a diagnostic result and reasoning path from the laboratory doctor's perspective. The radiology doctor analyzes the patient's medical imaging findings, providing a diagnostic result and reasoning path from the radiology doctor's perspective. The pathology doctor analyzes the patient's pathology findings, delivering a diagnostic result and reasoning path from the pathology doctor's perspective.

Since the agents analyze the data from their respective professional perspectives, the reasoning and clinical evidence from each agent may conflict with one another. To address this issue, we have designed a multi-agent cross-verification mechanism. When conflicts arise between agents, a discussion is initiated. Once a consensus is reached, each agent updates their evidence tree. After several rounds of discussion and merging, the final multi-agent evidence tree is produced.
Based on the final evidence tree, we can make a comprehensive diagnostic result, accompanied by the reasoning paths and all relevant clinical evidence. The pseudo code of workflow is shown in Algorithm~\ref{alg:multi_agent_medical}.

\begin{algorithm}[htbp]
\footnotesize
\caption{The pseudo code of our workflow.}
\label{alg:multi_agent_medical}
\KwIn{Domain agents $\{a_1, a_2, \dots, a_n\}$ (including Outpatient Agent, Laboratory Agent, Radiology Agent, Pathology Agent), initial medical data $D_0$, maximum rounds $k$, maximum turns per round $t$, prompts for discussion and final decision}
\KwOut{Final diagnosis tree $E_f$ and selected diagnosis option}

// Initialize variables\\
$round\_num \gets 0$\;

// Initialize diagnosis trees for each agent\\
$E_{cur} \gets \emptyset$\;
\For{$i \gets 1, \dots, n$}{
    $E_{a_i} \gets \text{AgentDiagnosis}\text{(}D_0, a_i\text{)}$; // Each agent generates its initial diagnosis tree based on $D_0$\\
    $E_{cur} \gets E_{cur} \cup \{E_{a_i}\}$\; 
}

// Multi-Agent Discussion\\
\While{$round\_num < k$}{
    $round\_num \gets round\_num \text{+} 1$\;
    
    \For{$turn\_num \gets 1, \dots, t$}{
        \For{$i \gets 1, \dots, n$}{
            \If{$\text{ShouldParticipate}\text{(}a_i, E_{cur}\text{)}$ is True}{
                $targets \gets \text{ChooseDiscussionTargets}\text{(}a_i, \{a_1, a_2, \dots, a_n\}\text{)}$\;
                \For{$j \in targets$}{
                    $opinion_{ij} \gets \text{GenerateOpinion}\text{(}a_i, a_j, E_{cur}\text{)}$\;
                    \text{LogInteraction}($a_i$, $a_j$, $opinion_{ij}$)\;
                }
            }
        }
    }
    
    // Update diagnosis trees based on discussion\\
    \For{$i \gets 1, \dots, n$}{
        $E_{a_i} \gets \text{UpdateDiagnosisTree}\text{(}E_{a_i}, \text{InteractionLog}\text{)}$\;
    }
}

// Final Decision\\
$E_f \gets \text{MakeFinalDecision}\text{(}\{E_{a_1}, E_{a_2}, \dots, E_{a_n}\}, \text{options}\text{)}$\;

\KwRet{$E_f$, selected diagnosis option}
\end{algorithm}

\section{Experiment}

\subsection{Experimental Setting}

\subsubsection{Dataset}
In this work, we construct a dataset based on real inpatient information from a large public tertiary hospital, which includes various types of medical data, including patient basic information (e.g., sex, age, present Illness), laboratory indicators (e.g., blood routine), radiology findings (e.g., X-ray, MRI) and pathology findings (e.g., frozen pathology).
This dataset contains complete examination data of 952 patients during hospitalization. The data covers 37 different clinical departments and involves 4,823 diseases. 
Please refer to appendix for the specific dataset construction process and detailed statistical information.

\subsubsection{Baseline Methods}


We compare our model with the following models:

(i) \textbf{Large Language Models.} We select several LLMs for comparison, including GPT-4o \cite{achiam2023gpt}, o1 \cite{jaech2024openai}, Claude 3.7 Sonnet \cite{TheC3} DeepSeek-V3 \cite{liu2024deepseek}, and DeepSeek-R1 \cite{guo2025deepseek}. These models not only perform excellently on general tasks but also rank among the top on medical domain benchmarks.

(ii) \textbf{Prompt-Based Methods.} This approach includes techniques like chain-of-thought (CoT) \cite{wei2022chain}, tree-of-thought (ToT) \cite{yao2023tree}, and MindMap \cite{wen2024mindmap}, which enhance model performance through optimized prompt design.

(iii) \textbf{Multi-Agent Systems.} Recently, several multi-agent methods tailored for the medical domain have been proposed, such as MedAgents \cite{tang2024medagents}, MDAgents \cite{kim2024mdagents}, and MMA \cite{peng2024integration}. These methods primarily focus on medical question answering tasks, where multiple LLMs collaborate to answer a single question.

For more detailed introduction of the above baselines, please see the appendix in supplementary material.

\begin{table*}[t]
\centering
\caption{Overall performance of different methods on the proposed dataset. \textbf{P.} means precision and  \textbf{R.} means recall. The best results are marked in \textbf{bold}.}
\label{tab:results}
\begin{tabular}{llccccc}
\toprule
\textbf{Category} & \textbf{Methods} & \textbf{P.(\%)} & \textbf{R.(\%)} & \textbf{F1(\%)} & \textbf{Relevance} & \textbf{Completeness}   \\
\midrule
\multirow{4}{*}{LLMs} & GPT-4o (Achiam et al. 2023) \cite{achiam2023gpt} & 88.89 & 29.58 & 44.39 & 3.57 & 3.2  \\
 & o1 (Jaech et al. 2024) \cite{jaech2024openai} & 87.63 & 27.61 & 42.00 & 3.6 & 3.03  \\
 & Claude 3.7 (Anthropic. 2025) \cite{TheC3} & 88.21 & 30.02 & 44.80 & 3.5 & 2.97  \\
 & DeepSeek-V3 (Liu et al. 2024) \cite{liu2024deepseek} & 89.01 & 31.59 & 46.63 & 3.43 & 3.2  \\
 & DeepSeek-R1 (Guo et al. 2025) \cite{guo2025deepseek} & 87.96 & 30.70 & 45.51 & 3.53 & 3.07   \\
\midrule
\multirow{4}{*}{Prompt-Based}
 & CoT (Wei et al. 2023) \cite{wei2022chain} & 90.32 & 34.04 & 49.45 & 4.0 & 3.47   \\
 & ToT (Yao et al. 2023) \cite{yao2023tree} & 79.23 & 28.13 & 41.52 & 4.2 & 3.3   \\
 & MindMap (Wen et al. 2024) \cite{wen2024mindmap} & 69.23 & 25.62 & 37.40 & 4.07 & 3.17   \\
\midrule
\multirow{3}{*}{Multi-Agent} 
  & MedAgents (Tang et al. 2024) \cite{tang2024medagents} & 92.08 & 33.52 & 49.15 & 4.4 & 3.37   \\
  & MDAgents (Kim et al. 2024) \cite{kim2024mdagents}  & 86.00 & 32.31 & 46.97 & 4.43 & 3.5   \\
  & MMA (Peng et al. 2024) \cite{peng2024integration} & 94.23 & 35.68 & 51.76 & \textbf{4.47} & 3.5   \\
\midrule
Our & Tree-of-Reasoning (ToR) & \textbf{95.70} & \textbf{46.60}  & \textbf{62.68} &  4.43 & \textbf{4.0}   \\
\bottomrule
\end{tabular}
\end{table*}

\subsubsection{Experimental Details}
For LLM methods, we utilize the APIs from corresponding platforms, including OpenAI\footnote{https://openai.com/}, Claude\footnote{https://claude.ai/}, DeepSeek\footnote{https://www.deepseek.com/}.
For multi-agent methods, prompt engineering, and our framework, we use DeepSeek-V3 as the fundamental LLM to ensure a fair comparison.
For MedRAG, we employ PubMed\footnote{https://pubmed.ncbi.nlm.nih.gov/}, StatPearls\footnote{https://www.statpearls.com/}, TextBooks, and Wikipedia\footnote{https://www.wikipedia.org/} as corpora, and BM25 \cite{robertson2009probabilistic} as the retriever. The top three most relevant results retrieved were used as context documents and incorporated into our framework.
We set maximum rounds $k$ = 2 and maximum turns per round $t$ = 2 in our framework. 
All experiments were conducted in a zero-shot setting.

\subsubsection{Evaluation Metrics}

To provide an objective evaluation metric for all methods, we define the task as a multi-label classification problem. Specifically, we consider all the diagnosed diseases of a single patient as the correct labels, while randomly selecting other diseases from the same department as the incorrect labels. This allows us to generate both correct and incorrect labels for each sample.
To assess the model's performance on this task, we use \textbf{precision}, \textbf{recall}, and \textbf{F1} score as evaluation metrics.
In addition, we introduce subjective evaluation criteria: \textbf{relevance} and \textbf{completeness}. We select three real-world doctors to score the diagnostic explanations provided by all methods in order to evaluate their value in real-world clinical diagnosis. 
Relevance refers to whether the output is related to the case without irrelevant or redundant information. 
Completeness assesses whether the output provides all the key evidences needed to make a diagnosis with no omissions.
Each of these dimensions is scored on a scale from 0 to 5, and is precised to one decimal place. 
For more details of the definition of metrics, please refer to the appendix in supplementary material.

\subsection{Main Results}

\begin{table*}[t]
\centering
\caption{Ablation study on different Agents. ``$\checkmark$'' means presence of the doctor agent and ``$\times$'' means absence of the doctor agent.}
\label{tab:abla_role}
\begin{tabular}{ccccccc}
\toprule
\textbf{Outpatient Doctor} & \textbf{Laboratory Doctor} & \textbf{Radiology Doctor} & \textbf{Pathology Doctor} & \textbf{P.} & \textbf{R.} & \textbf{F1} \\
\midrule
$\checkmark$ & $\times$ & $\times$ & $\times$ & 89.01 & 31.59 & 46.63 \\
$\checkmark$ & $\checkmark$ & $\times$ & $\times$ & 90.12  & 33.98 & 49.35 \\
$\checkmark$ & $\checkmark$ & $\checkmark$ & $\times$ & 94.07 & 44.67 & 60.58 \\
$\checkmark$ & $\checkmark$ & $\checkmark$ & $\checkmark$ & \textbf{95.70} & \textbf{46.60}  & \textbf{62.68}   \\
\bottomrule
\end{tabular}
\end{table*}

\begin{table}[t]
\centering
\caption{Ablation study on different mechanism or module. ``w/o'' means ToR framework without a certain part.}
\label{tab:abla_module}
\begin{tabular}{cccccc}
\toprule
\textbf{Abalation Model} & \textbf{P.} & \textbf{R.} & \textbf{F1}\\
\midrule
ToR w/o Evidence Tree &  84.87 & 36.52  &  51.07 \\
ToR w/o Cross Verification & 90.33  & 41.84  & 57.19  \\
ToR w/o MedRAG & 93.07 & 44.21 & 59.94 \\
ToR  & \textbf{95.70} & \textbf{46.60} & \textbf{62.68}  \\
\bottomrule
\end{tabular}
\end{table}

As shown in Table~\ref{tab:results}, we compare the performance of the LLM-based methods, the prompt-based methods, the multi-agent methods, and our proposed framework using objective evaluation metrics (Precision, Recall, and F1) and subjective evaluation metrics (Relevance and Completeness). We identify the following key findings:

(i) The first finding is that multi-agent methods almost outperform both LLM-based and prompt-based methods. Specifically, the F1 score of MMA is 51.76\%, while the F1 scores for DeepSeek-R1 and CoT are 45.51\% and 49.45\%, respectively. 
Unlike traditional medical diagnosis tasks, the complex medical diagnosis tasks in this work involve multiple types of medical data, requiring various medical tasks to achieve consistent diagnostic results. A single LLM approach must handle input from different tasks simultaneously, which may cause the model’s attention to be scattered and unable to effectively focus on the key information of each task, ultimately affecting overall performance. In contrast, the multi-agent system can break down the complex medical tasks into smaller sub-tasks, with each agent focusing on its respective task without interference from other medical data. By collaborating, the agents provide consistent diagnostic results, enabling the system to capture more clinical evidence and produce a more comprehensive diagnosis.

(ii) The second finding is that our framework outperforms existing multi-agent methods in objective metrics. Specifically, the F1 score of our framework is 62.68\%, compared to 51.76\% for MMA. Unlike existing multi-agent methods, our method explicitly presents the clinical reasoning path with evidence tree following ``evidence-based medicine'' theory. This allows agents to re-examine the diagnosis from their specialized perspectives and correct any conflicting or missing evidence in this tree. This structured exchange of information between reasoning paths enables our method to consider the patient's disease more comprehensively, leading to improved diagnostic results.

(iii) The third finding is that our method’s subjective evaluation metrics of relevance is comparable to those of other methods, while the completeness metric is better than other methods. Specifically, our method gain 0.5 score improvement compared with MMA method. This indicates that the experts who conducted the evaluation believe that our methods can explore more clinical evidence to support diagnostic views. In real-world medical settings, this could provide doctors with more comprehensive clinical references, making it highly significant for clinical decision support systems.

\begin{figure}[t]
  \centering
  \includegraphics[width=0.8\linewidth]{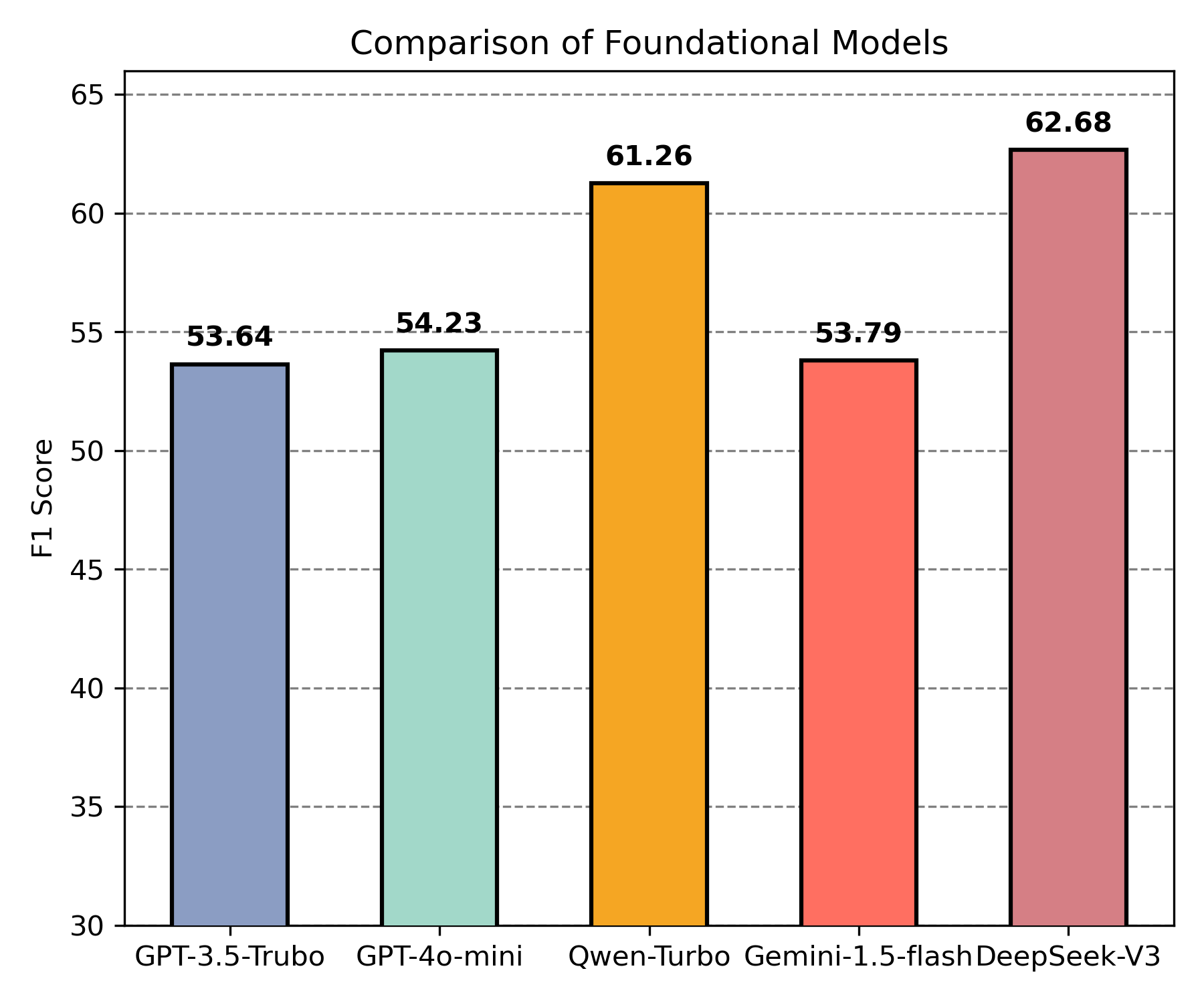}
  \caption{Comparison of Different Foundational Models in the proposed ToR framework.}
  \label{fig:foundational_model}
\end{figure}

\subsection{Ablation Study}


To investigate the impact of different doctor agents on final results, we conduct an ablation study, as shown in Table~\ref{tab:abla_role}. This experiment selectively excludes specific doctors and observes the changes in Precision, Recall, and F1, indicated by the presence ($\checkmark$) or absence ($\times$) of roles in the system configuration. When a specific doctor agent is excluded, the corresponding medical data is reassigned to the outpatient doctor agent.
We make several key findings:

(i) As the number of agents increases, the model's performance improves. Specifically, when the laboratory doctor was added, the F1 score increased by 2.72\%. Further improvements were observed when the radiology and pathology doctors were added, with F1 scores increasing by 11.23\% and 2.10\%, respectively. A closer examination reveals that the increase in F1 score primarily comes from an improvement in Recall. This suggests that as the number of agents increases, the model is able to consider a more comprehensive analysis of different types of medical data, improving disease coverage while maintaining diagnostic accuracy.

(ii) The inclusion of the radiology doctor has the most significant impact on performance. Specifically, when the radiology doctor was added, the F1 score increased by 11.23\%, indicating that the analysis of medical imaging findings is crucial for complex medical diagnosis tasks. Medical imaging data encompasses the majority of tumor-related diseases, which require specialized knowledge in radiology for accurate diagnosis. The introduction of the radiology doctor allows for a focused clinical reasoning approach to these diseases, enhancing the overall diagnostic coverage.

(iii) When the laboratory, radiology, and pathology doctors are all excluded, the entire multi-agent framework degrades to a single LLM, resulting in the lowest F1 score of 46.63\%. When handling complex medical diagnosis task, a single LLM must simultaneously process multiple types of medical data, which can cause the model’s attention to shift and prevent it from focusing on the key information from each type of data. This leads to omissions and, consequently, a decline in diagnostic performance.

To explore the impact of different mechanism or module on the overall framework, we conduct corresponding ablation experiments, as shown in Table~\ref{tab:abla_module}.
``w/o Evidence Tree'' refers to the scenario where the outputs between agents are directly presented as unstructured natural language. In this case, the F1 score decreased by 11.61\%. The medicine evidence tree explicitly preserves the reasoning paths and clinical evidence of different agents in complex medical reasoning scenarios. This helps prevent evidence drift or omission during multi-agent interactions, thereby improving the overall clinical collaboration and reasoning capability.
The addition of the multi-agent cross-verification mechanism allows agents to re-examine their diagnostic processes and clinical evidence from different perspectives. This enables the correction of contentious reasoning steps, resulting in an increase in F1, specifically improving by 5.49\%.
MedRAG provides domain knowledge to different expert agents, enhancing the diagnostic performance of a single LLM and consequently improving the overall system's ability. Specifically, after incorporating the MedRAG tool, the F1 score increased by 2.74\%.

\begin{figure*}[h]
  \centering
  \includegraphics[width=0.85\linewidth]{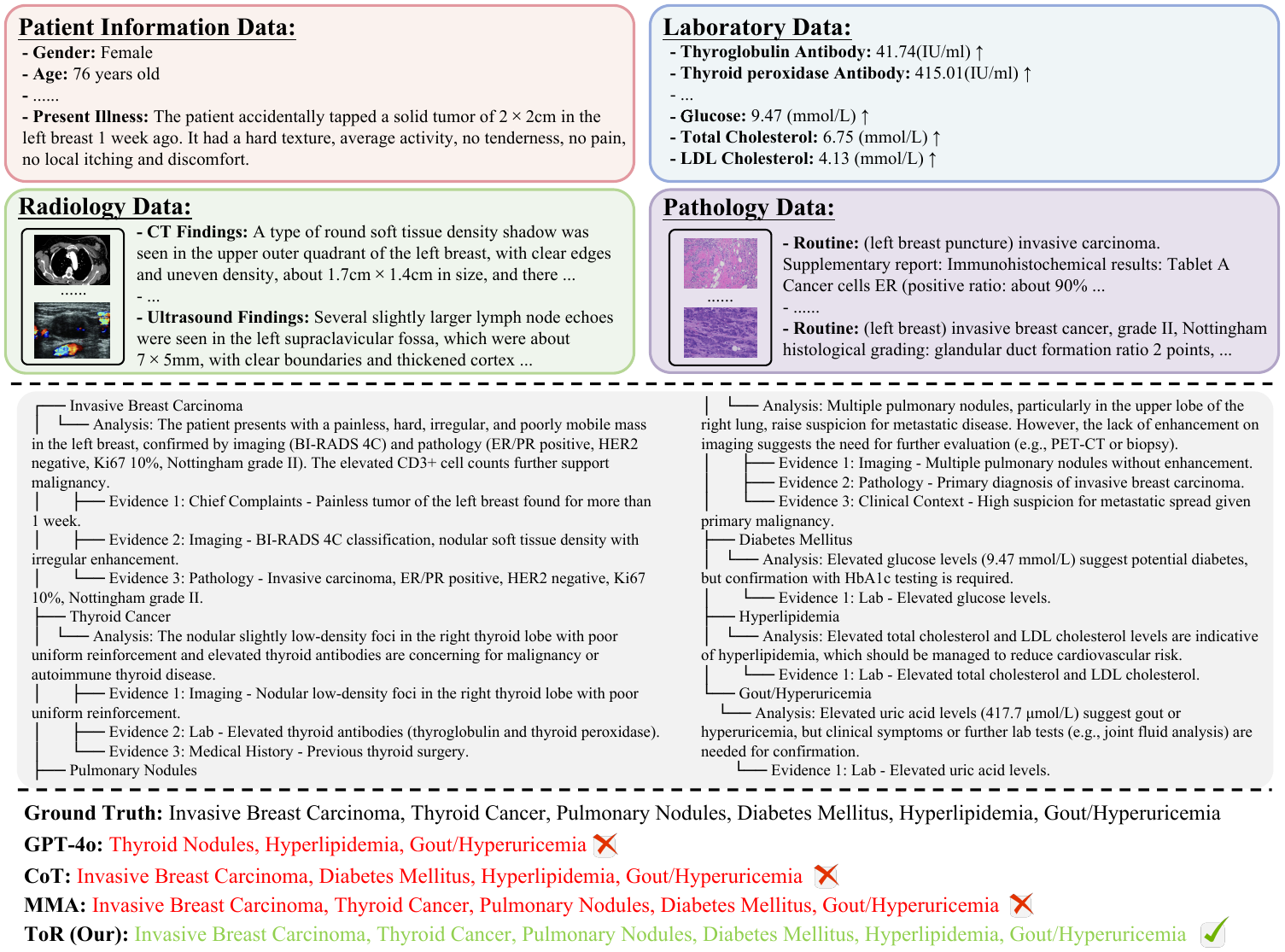}
  \caption{Case study for a given patient case. The top part is the different type of medical data of the patient. The middle part is the reasoning output with the evidence tree of our ToR framework. The bottom part is the ground truth and different answers of GPT-4o, CoT, MMA and ToR.}
  \label{fig:case}
\end{figure*}

\subsection{Comparison of Different Foundational Models}

To test the performance of the ToR framework under different foundation models, we experiment with five cost-effective models, including GPT-3.5-Turbo, GPT-4o-mini, Qwen-Turbo, Gemini-1.5-flash, and DeepSeek-V3. Among these models, GPT-3.5-Turbo has the highest cost of \$1.50 per million tokens, while Qwen-Turbo has the lowest cost of \$0.6 per million tokens. The experimental results are shown in Figure ~\ref{fig:foundational_model}, which shows the F1 value of the ToR method under different foundation models. The results show that when the foundation model is DeepSeek-V3, ToR has the best performance, with an F1 value of 62.68\%. When the foundation model is GPT-3.5-Turbo, ToR has the worst performance, with an F1 value of 53.64\%. Taking into account both cost and performance, we finally choose DeepSeek-V3 as the foundational model of our framework.

\subsection{Case Study}


Figure~\ref{fig:case} shows a case study of a complex medical diagnosis scenario. The top of the figure shows all the medical examination data of real patients, which covers four types: patient information, laboratory medical indicators, radiology findings, and pathology findings. The middle part of the figure shows the process of multi-agent collaboration in our proposed ToR framework, including explicit reasoning paths and evidence trees for different diseases. The bottom of the figure lists the diagnosis results of different methods, including GPT-4o (LLM-based method), CoT (prompt-based method), MMA (multi-agent method) and our framework. The results show that GPT-4o mistakenly diagnosed Thyroid Cancer as Thyroid Nodules. At the same time, GPT-4o, CoT and MMA all missed the diagnosis of Pulmonary Nodules. This shows that when processing a large amount of medical data at the same time, the limited reasoning depth of these methods makes it difficult to capture key evidence information, resulting in reasoning errors or omissions. In contrast, the ToR method considers all possible diseases through explicit reasoning based on the evidence tree and finally makes a correct diagnosis. This shows that our method has stronger performance in complex task reasoning scenarios.

\section{Conclusion}
In this study, we propose Tree-of-Reasoning (ToR), a multi-agent framework designed to handle complex medical diagnostic task by introducing a novel reasoning tree structure based on clinical evidence. 
Our method first assigns different domain-specific doctor agents to analyze various types of medical data and generate corresponding reasoning paths and evidence. Then we introduce a multi-agent cross-verification mechanism to update the reasoning process and corresponding evidence of each agent, leading to a final multi-agent collaborative diagnosis result. This explicit multi-agent reasoning framework not only enhances the reasoning capability in complex scenarios but also provides better interpretability. Experiments show that the ToR method outperforms other baselines, highlighting the benefits of incorporating reasoning trees to improve medical diagnosis.

\begin{acks}
This research is supported by the National Natural Science Foundation of China (62476097), the Fundamental Research Funds for the Central Universities, South China University of Technology (x2rjD2250190),  Guangdong Provincial Fund for Basic and Applied Basic Research—Regional Joint Fund Project (Key Project) (2023B1515120078), Guangdong Provincial Natural Science Foundation for Outstanding Youth Team Project (2024B1515040010), the Hong Kong Polytechnic University under the Postdoc Matching Fund Scheme (Project No. P0049003).
\end{acks}

\bibliographystyle{ACM-Reference-Format}
\balance
\bibliography{sample-base}

\clearpage

\appendix

\section{Dataset Information}
\label{section:data_info}
\subsection{Dataset Construction}

\begin{figure}[h]
  \centering
  \includegraphics[width=\linewidth]{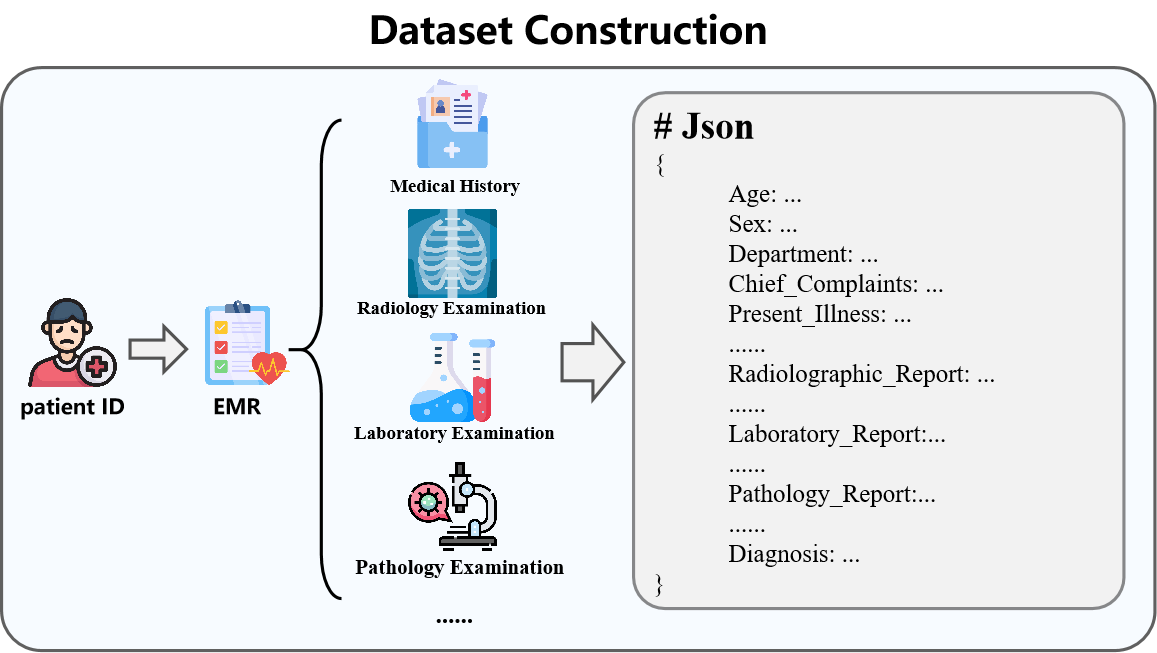}
  \caption{The construction process of our dataset.}
  \label{fig:data_construction}
\end{figure}

In this study, we construct a comprehensive dataset using real inpatient data from a large public tertiary hospital. The dataset construction process involves the following steps:

(i) Extraction of Patient Information: Unique patient IDs are first extracted from inpatient records. Using these IDs, patient-specific data is retrieved, and any privacy-sensitive information is anonymized to ensure confidentiality.

(ii) Accessing Electronic Medical Records (EMRs): Patient IDs are then used to access the hospital's EMR system. Detailed medical histories, including patient complaints, present illnesses, laboratory test results, and radiology examination reports, are extracted directly from the EMRs.

The final dataset consists of records from a total of 952 patients during hospitalization, spanning multiple medical departments, including, but not limited to, oncology, gastrointestinal surgery  and thyroid surgery.

Through the process outlined above, we obtain multi-source medical data that offers a rich diversity of medical information. This includes: 

(1) Comprehensive Medical Histories: From initial complaints to detailed physical examination findings, providing a holistic view of each patient's condition. 

(2) Diverse Examination Data: Such as laboratory test reports, medical imaging reports, and pathology reports. 

(3) In-Depth Clinical Documentation: Encompassing a wide range of medical procedures performed across various departments.

The collected data is structured and stored as JSON files, with a specific case in Table~\ref{tab:app_case}. The overall data collection process is illustrated in Figure~\ref{fig:data_construction}.

\subsection{Detailed statistics of the dataset}

\begin{figure*}[h]
  \centering
  \includegraphics[width=\linewidth]{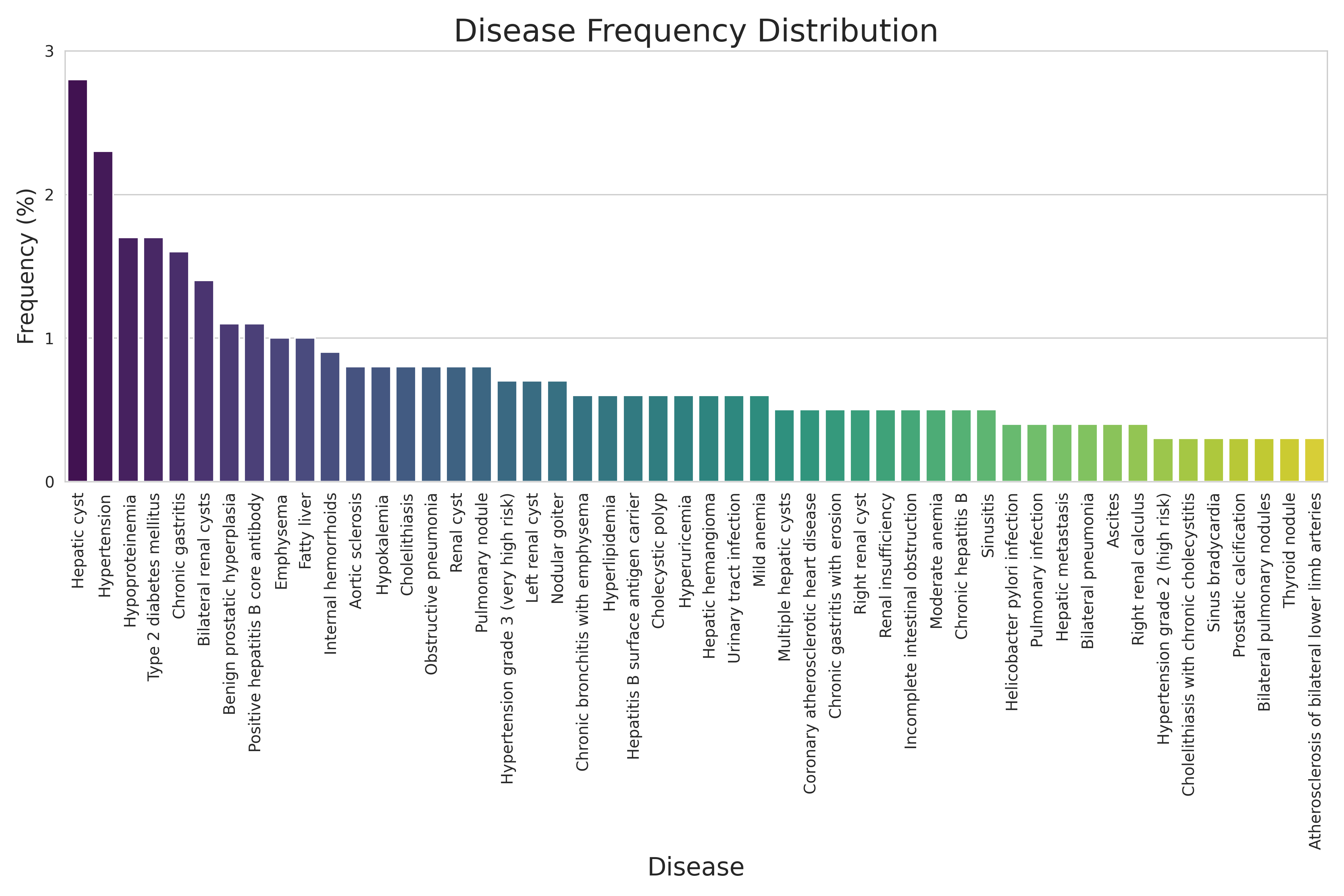}
  \caption{The statistics of labels. The horizontal ordinate represents the distribution of disease labels, and the vertical represents the corresponding frequency in the dataset.}
  \label{fig:disease_freq}
\end{figure*}

The disease statistics of our dataset are shown in Figure~\ref{fig:disease_freq}, with 4,823 diseases in total. Due to space limitations, the figure only shows the distribution of some labels from high to low. Among them, the top five labels are Hepatic cyst (2.8\%), Hypertension (2.3\%), Hypoproteinemia (1.7\%), Type 2 diabetes mellitus (1.7\%) and Chronic gastritis (1.6\%). 

\begin{figure*}[]
  \centering
  \includegraphics[width=0.8\linewidth]{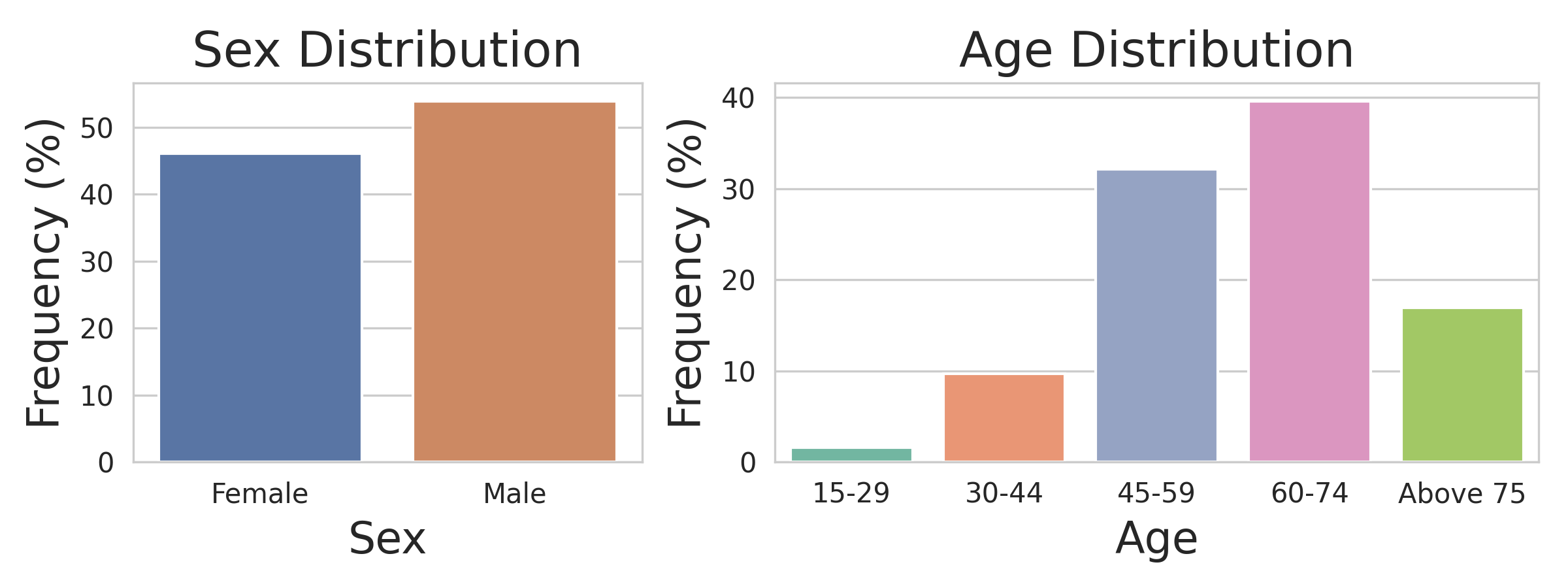}
  \caption{The statistics of patient information. The left image represents the gender distribution, and the right image represents the distribution of different age groups.} 
  \label{fig:patient_info_distribution}
\end{figure*}

The patient statistics are shown in Figure~\ref{fig:patient_info_distribution}, among which female patients account for 46.1\% and male patients account for 53.9\%. As for the age, most patients are young to middle-aged, among which patients aged 60-74 account for 39.6\%, and elderly patients (over 75 years old) account for nearly 16.9\%.

\begin{figure*}[h]
  \centering
  \includegraphics[width=\linewidth]{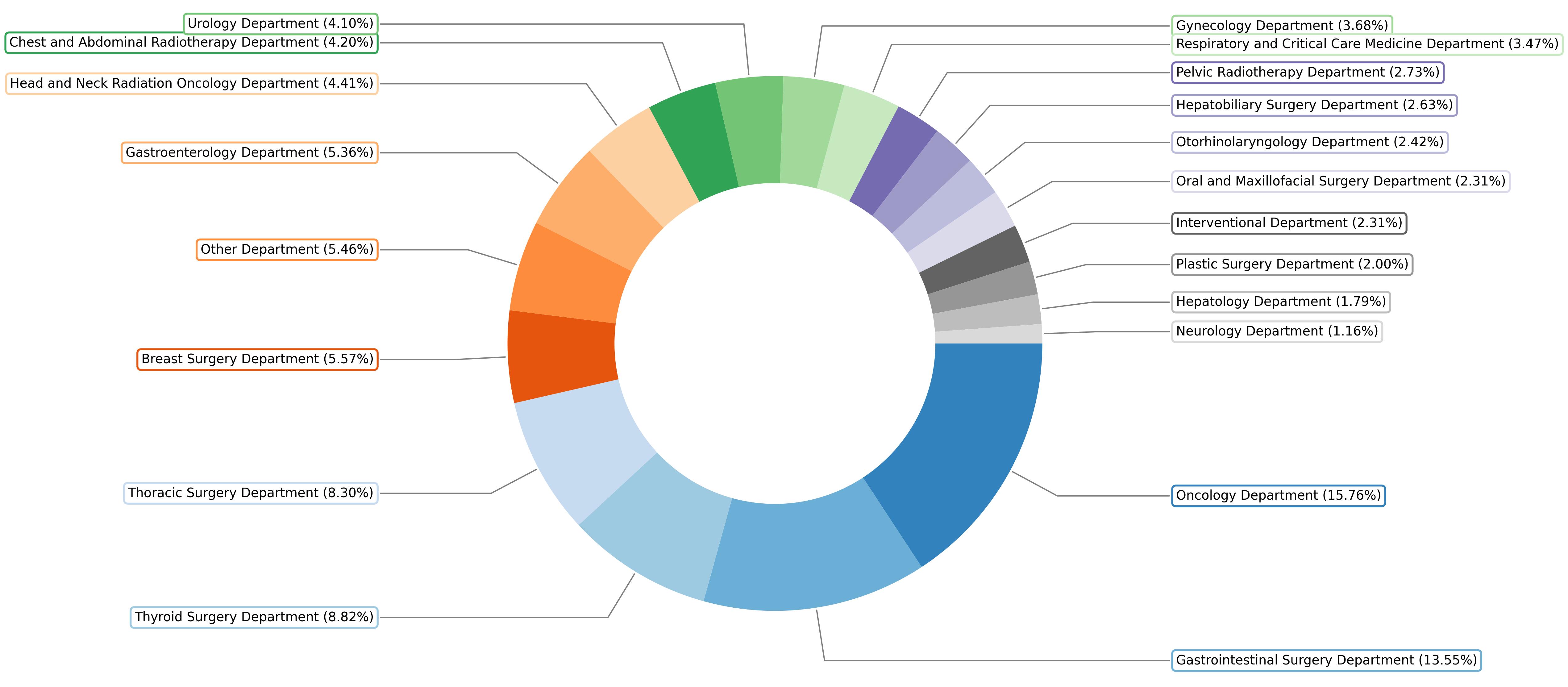}
  \caption{The statistics of departments in dataset.}
  \label{fig:department}
\end{figure*}

The department statistics are shown in Figure~\ref{fig:department}, with 37 departments in total. The top three departments were oncology (15.76\%), gastrointestinal surgery (13.55\%), and thyroid surgery (8.82\%).

\section{Prompt}
\label{appendix:prompt}
Prompt templates of our method are presented in Table~\ref{tab:prompt1}-\ref{tab:prompt_final_diagnosis}.

\section{Experimental Settings}
\subsection{Baselines}
\label{sec:baselines}

\begin{itemize}[itemsep=2pt,topsep=0pt,parsep=0pt,leftmargin=0.5cm]

\item \textbf{GPT-4o} \cite{achiam2023gpt}: GPT-4o is a multilingual, multimodal generative pre-trained transformer developed by OpenAI and released in May 2024, achieving state-of-the-art results in voice, multilingual, and vision benchmarks.

\item \textbf{o1} \cite{jaech2024openai}: o1 is a new large language model trained with reinforcement learning to perform complex reasoning. o1 thinks before it answers—it can produce a long internal chain of thought before responding to the user.

\item \textbf{Claude 3.7 Sonnet} \cite{TheC3}: Claude 3.7 Sonnet was released on February 24, 2025. It is a pioneering hybrid AI reasoning model that allows users to choose between rapid responses and more thoughtful, step-by-step reasoning. This model integrates both capabilities into a single framework, eliminating the need for multiple models. Users can control how long the model "thinks" about a question, balancing speed and accuracy based on their needs.

\item \textbf{DeepSeek-V3} \cite{liu2024deepseek}: DeepSeek-V3 is a strong Mixture-of-Experts (MoE) language model with 671B total parameters with 37B activated for each token. To achieve efficient inference and cost-effective training, DeepSeek-V3 adopts Multi-head Latent Attention (MLA) and DeepSeekMoE architectures, which were thoroughly validated in DeepSeek-V2. Furthermore, DeepSeek-V3 pioneers an auxiliary-loss-free strategy for load balancing and sets a multi-token prediction training objective for stronger performance.

\item \textbf{DeepSeek-R1} \cite{guo2025deepseek}: DeepSeek-R1, which incorporates cold-start data before RL. DeepSeek-R1 achieves performance comparable to OpenAI-o1 across math, code, and reasoning tasks.

\item \textbf{chain-of-thought (CoT)} \cite{wei2022chain}: Chain-of-thought prompting is an approach in artificial intelligence that simulates human-like reasoning processes by delineating complex tasks into a sequence of logical steps towards a final resolution. This methodology reflects a fundamental aspect of human intelligence, offering a structured mechanism for problem-solving. In other words, CoT is predicated on the cognitive strategy of breaking down elaborate problems into manageable, intermediate thoughts that sequentially lead to a conclusive answer.

\item tree-of-thought (ToT) \cite{yao2023tree}: Tree of thoughts (ToT) is a ground-breaking framework designed to enhance the reasoning capabilities of large language models (LLMs). This approach simulates human cognitive strategies for problem-solving, enabling LLMs to explore multiple potential solutions in a structured manner, akin to a tree's branching paths.

\item \textbf{MindMap} \cite{wen2024mindmap}: MindMap leverages knowledge graphs (KGs) to enhance LLMs’ inference and transparency, enabling LLMs to comprehend KG inputs and infer with a combination of implicit and external knowledge.

\item \textbf{MedAgents} \cite{tang2024medagents}: MedAgents is a novel Multi-disciplinary Collaboration (MC) framework for the medical domain that leverages LLM-based agents in a role-playing setting that participate in a collaborative multi-round discussion, thereby enhancing LLM proficiency and reasoning capabilities.

\item \textbf{MDAgents} \cite{kim2024mdagents}: MDAgents, a framework designed to enhance the utility of LLMs in complex medical decision-making by dynamically structuring effective collaboration models. To reflect the nuanced consultation aspects in clinical settings, MDAgents adaptively assigns LLMs either to roles independently or within groups, depending on the task’s complexity. This emulation of real-world medical decision processes has been comprehensively evaluated, with MDAgents outperforming previous solo and group methods in 7 out of 10 medical benchmarks.

\item \textbf{MMA} \cite{peng2024integration}: MMA system is the first QA system to utilize LLM to simulate different medical specialists for collaborative diagnostic decision-making. The strength of the MMA system lies in its ability to assign different expert roles to handle various medical tasks, achieving joint decision-making through structured outputs and a shared information pool.


\end{itemize}

\subsection{Metrics}
\label{sec:metric}
\begin{itemize}[itemsep=2pt,topsep=0pt,parsep=0pt,leftmargin=0.5cm]

\item \textbf{Precision}: Precision measures the proportion of true positive predictions among all instances predicted as positive, calculated as:
\[
\text{Precision} = \frac{TP}{TP \text{+} FP}
\]
where TP is the number of true positives, and FP is the number of false positives.

\item \textbf{Recall}: Recall measures the proportion of actual positive instances correctly predicted by the model, calculated as:
\[
\text{Recall} = \frac{TP}{TP \text{+} FN}
\]
where TP is the number of true positives, and FN is the number of false negatives.

\item \textbf{F1}: The F1 score is the harmonic mean of precision and recall, commonly used to evaluate performance on imbalanced datasets, calculated as:
\[
\text{F1} = 2 \times \frac{\text{Precision} \times \text{Recall}}{\text{Precision} \text{+} \text{Recall}}
\]



\item \textbf{Relevance}:
\begin{itemize}
    \item \textbf{Definition:} The degree of direct connection and importance between the diagnostic results, analysis content, and supporting evidence provided by the model and the specific case being evaluated or the medical question being asked. The focus of the evaluation is on whether the model focuses on key information and avoids providing irrelevant or secondary content.
    \item \textbf{Scoring Guidelines and Requirements (0-5 points, precise to one decimal place):}
    \begin{itemize}[leftmargin=0.35cm]
        \item \textbf{5.0 points:} Completely relevant. The diagnostic results, analysis, and evidence provided by the model are highly relevant to the current case, directly answer the core questions, and do not introduce any irrelevant information. All information is crucial for understanding the condition and making decisions.
        \item \textbf{4.0 - 4.9 points:} Highly relevant. The vast majority of the information provided by the model is relevant to the current case, with only a very small amount of negligible non-core information. Overall, it has significant value for understanding the condition and making decisions.
        \item \textbf{3.0 - 3.9 points:} Moderately relevant. Most of the information provided by the model is relevant to the current case, but it may contain some non-directly related secondary information, or the relevance may be slightly weaker in some key aspects. Manual further screening and focusing are needed.
        \item \textbf{2.0 - 2.9 points:} Low relevance. Only a portion of the information provided by the model is directly relevant to the current case, containing more irrelevant or indirectly related information. Extensive manual screening is needed to find useful content.
        \item \textbf{1.0 - 1.9 points:} Very low relevance. The relevance of the information provided by the model to the current case is very weak, with most of the content unrelated to the question being evaluated.
        \item \textbf{0.0 - 0.9 points:} Completely irrelevant. The diagnostic results, analysis, and evidence provided by the model are completely unrelated to the current case.
    \end{itemize}
\end{itemize}

\item \textbf{Completeness}:
\begin{itemize}
    \item \textbf{Definition:} Whether the diagnostic results, analysis content, and supporting evidence provided by the model contain all the key elements and information necessary for making reasonable judgments and further decisions. The focus of the evaluation is on whether the model has omitted important considerations, data, or explanations.
    \item \textbf{Scoring Guidelines and Requirements (0-5 points, precise to one decimal place):}
    \begin{itemize}[leftmargin=0.35cm]
        \item \textbf{5.0 points:} Completely complete. The model provides all the key information needed for comprehensive judgment and decision-making, without omitting any important diagnostic elements, analysis steps, or supporting evidence.
        \item \textbf{4.0 - 4.9 points:} Highly complete. The model provides the vast majority of key information, with only some minor omissions that do not affect the core judgment. Overall, it can support relatively comprehensive understanding and decision-making.
        \item \textbf{3.0 - 3.9 points:} Moderately complete. The model provides some key information, but may have omitted some important diagnostic elements, analysis steps, or supporting evidence, requiring manual supplementation to make a more comprehensive judgment.
        \item \textbf{2.0 - 2.9 points:} Low completeness. The information provided by the model is insufficient to make a reasonable judgment, with many key diagnostic elements, analysis steps, or supporting evidence missing, requiring extensive manual supplementation of information.
        \item \textbf{1.0 - 1.9 points:} Very low completeness. The information provided by the model is very limited and contains almost no core elements needed for judgment.
        \item \textbf{0.0 - 0.9 points:} Completely incomplete. The model does not provide any meaningful diagnostic results, analysis, or evidence.
    \end{itemize}
\end{itemize}

\end{itemize}

\clearpage

\clearpage

\begin{onecolumn}
\begin{longtable}{p{\linewidth}}
\caption{A case in the dataset.}
\label{tab:app_case}\\
\toprule
\endfirsthead
\multicolumn{1}{c}{{\tablename\ \thetable{} -- Continued from previous page}} \\
\toprule
\endhead
\midrule
\multicolumn{1}{r}{{Continued on next page}} \\
\endfoot
\bottomrule
\endlastfoot

\textbf{"Age":} "Y70",\\
\textbf{"Sex":} "male",\\
\textbf{"Chief-Complaints":} "Continuous hoarseness for more than 2 months.",\\
\textbf{"Present-Illness":} "The patient had hoarseness without obvious causes 2 months ago. It was intermittent and then persistent. He spoke more difficult, no sore throat, fever, no cough, sputum, no afternoon hot flashes, no progressive swallowing or difficulty breathing. He had been treated in a local hospital (not detailed) with medication (specific details) and had no relief. He had mild shortness of breath in the past month and was obviously inhaled. He had been to the local hospital for treatment, but the effect was poor and gradually worsened. Now he went to our hospital for further treatment. The outpatient electronic nasopharyngeal and laryngoscopy showed a \"throat tumor\", and now he was hospitalized. Since the onset of the disease, the patient's mental reaction was average, his stomach was still acceptable, and there was no abnormality in urination and defecation.",\\
\textbf{"Physical-Examination":} "T:36℃, P: 91 times/min, R: 20 times/min, BP: 134/92mmHg. Ear: There is no deformity or traction pain in the auricle, the double external auditory canal is unobstructed, the tympanic membrane is complete, the light cone is obvious, and there is no tenderness in the mastoid area. Nose: There is no deformity in the outer nose, no wings are trembling, the lower turbinates are slightly thick, no new organisms or abnormal secretions are seen in the middle nasal passages, and no obvious deviation is seen in the nasal septum. Throat: The throat is slightly congested, the double tonsils are hypertrophy, the surface is congested, the epiglottis can be lifted, the lymphatic follicles at the base of the tongue are hyperplasia, the bilateral cycloallium joints are slightly swollen, the bilateral vocal cords are slightly edema, new organisms are seen in the anterior middle of the left vocal cord, there are more viscous white membranes on the surface, no erosion and ulcers are seen, and the glottal closure is poor.",\\
\textbf{"Laboratory-Examination":}"\{"Routine blood, blood biochemistry": "basically normal", "squamous epithelial cell carcinoma antigen": "0.40(ng/ml)"\}",\\
\textbf{"Ultrasonic-Imaging-Examination":} "Ultrasound imaging examination 1: The vascular morphology and structure of the arteries at all levels of the lower limbs (bilateral common femoral artery, superficial femoral artery, popliteal artery, anterior tibial artery, posterior tibial artery, dorsal foot artery) of both sides is still normal, the inner diameter of the vascular lumen is normal, and the echo of the inner middle membrane has no obvious thickening. A small number of scattered and distributed sclerotic spots can be seen. Color Doppler shows that the blood flow in the lumen is continuous and complete, the margins are regular, and the blood flow spectrum is normal. The vascular structure of the bilateral lower limb veins (bilateral common femoral vein, superficial femoral vein, popliteal vein, posterior tibial vein, and great saphenous vein) is normal, the wall of the tube is not thick, the endometrium is smooth, and no echo of the thrombus clump is seen in it. Color Doppler shows that the blood flow in the lumen is continuous and intact, and the venous return is normal. Ultrasound imaging examination 2: The liver section of the part of the liver is not clearly displayed due to intestinal qi interference. The liver section was seen with normal morphology and flat surface. A dark area of cystic fluid can be seen in the left liver, with a thin wall, clear boundary, clear capsule fluid, and enhanced echo at the posterior. The first half of the remaining intrahepatic echo is fine and enhanced, the back half is sparse and attenuated, and the distribution is still uniform. The structure of the intrahepatic duct can still be displayed, with no thickening and no expansion of the portal vein trunk. The gallbladder was filled with poor amounts and no abnormal echoes were seen inside. There was no dilation of the intrahepatic bile duct and no stones; there was no dilation of the common bile duct and no abnormalities were found in it. The spleen is of normal shape and size, with small and even echoes, and no abnormalities are seen inside. The portal vein and splenic vein are smooth, without dilation, no embolism is seen, the colored blood flow is good, and the flow rate is normal. The pancreas is of normal shape and size, with uniform echoes, no tumors are seen, and the pancreatic duct is not expanded, and no obvious abnormalities are seen in it.  The left kidney is in normal shape and size, and the renal parenchymal echoes have not been enhanced. A strong echo spot can be seen in the lower pot, with a size of about 7×6mm, accompanied by sound shadows at the back, and no tumors or effusions are seen in the kidney. The right kidney has normal morphology and size, but the renal parenchymal echoes have not been enhanced, the light spot group of the assembly system is evenly distributed, and no tumors, stones or effusions are found in the kidney. Color Doppler imaging: The blood flow signal in both kidneys is rich, and the renal vascular tree structure is normal. There was no dilation of the bilateral ureter and no obvious abnormal echoes were seen. The bladder was filled with poor, and no tumors or stone echoes were seen inside. The prostate has normal morphology and size, complete envelope, symmetrical appearance, normal proportion of internal and external glands, strong echo light spots can be seen inside, and no obvious place-occupying echoes are seen.",\\
\textbf{"X-ray-Imaging-Examination":} "X-ray imaging 1: The texture of the bilateral lung field was slightly thickened, and no clear parenchymal lesions were seen; no enlargement or thickening of the bilateral hilar, and no widening of the mediastinum; the size, shape and position of the heart shadow were normal; arc-shaped calcified shadows could be seen in the aortic arch; the bilateral diaphragm surface was bright and the bilateral diaphragm angle was sharp; no clear abnormalities were seen in the ribs indicated.",\\
\textbf{"CT-Imaging-Examination":} "CT imaging examination 1: The throat structure was clear, no abnormal density was seen, and no stenosis was seen on both sides of the piriform fossa; the bilateral vocal cords were irregularly thickened, with a thicker area of about 0.9cm, obvious on the left, combined with anterior joint involvement, and bone destruction of adjacent thyroid cartilage. There was no abnormality in the morphology, size and density of the bilateral thyroid gland, and no abnormal strengthening was seen after enhancement. Several small lymph nodes can be seen on both sides of the neck. The larger ones have a short diameter of about 0.6cm, and the boundary is clear, and the reinforcement is evenly enhanced. The structure of the bilateral neck muscle group was symmetrical, and no abnormal changes were seen. CT imaging 2: Three-dimensional stereotactic navigation after large aperture CT radiotheMagnetic resonance imaging examinationrapy.",\\
\textbf{"Magnetic-Resonance-Imaging-Examination":} "Magnetic resonance imaging examination 1: uneven thickening of bilateral vocal cords ~ front joint (point on the left), T1WI presents an equal signal, T2WI presents a slightly higher signal, DWI sequence observation is under clear, enhancement scan shows under even strengthening, the larger axis range is about 1.6cm×0.7cm, the boundary is under clear, corresponding laryngeal cavity becomes narrower, and the bilateral pyriform fossa becomes slightly narrower. There were no clear abnormalities in thyroid cartilage, bilateral arytenoid cartilage, and cyclic cartilage. After tracheotomy, the soft tissue around the surgery area slightly swelled and local exudation changed. There are multiple small and slightly larger lymph nodes in the bilateral neck areas I to V, with clear boundaries. The larger ones have a short diameter of about 0.6cm, which enhances the scanning and strengthening more evenly. The bilateral palatal tonsils and lingual tonsils are swollen, the signal and strengthening are still uniform, and the pharyngeal cavity becomes narrower accordingly. There were no obvious abnormalities in the bilateral parotid and submandibular gland morphology, size, signal and enhancement. There are no clear abnormalities in the morphology, size and signal of the bilateral lobes and isthmus of the thyroid gland, and no clear abnormal enhancement is seen. There was no stenosis in the nasopharyngeal cavity, no significant thickening of the nasopharyngeal mucosa, no obvious tumors were seen, no abnormal strengthening was seen in the enhancement scan, and no bilateral pharyngeal crypts were seen to become shallow. There were no obvious abnormalities in the bilateral palatine tendonalis muscle, palatine levator, cephalos muscle, intraphic muscle, and phthaloid plate. No significant enlarged lymph nodes were seen after both pharynx. There was no abnormality in the bilateral phalerial fossa and bilateral masticatory muscle space. There were no obvious abnormalities in the skull base slope, sphenoid basal base, bilateral rock apex, and occipital basal bones, and no obvious abnormalities were found in the bilateral rupture hole and next to the sublingual nerve canal. The bilateral cavernous sinus morphology was normal and there was no widening. The mucosa of some sinus ethmoid sinus is slightly thickened, no obvious secretions are found on both sides of the mastoid process, and the bone wall is intact. The size, shape and position of the eyes were normal, no abnormal signals or reinforcement shadows were seen, and no placeholder lesions were seen behind the ball. The morphology of the butterfly is normal, and no placeholder lesions are found in it.",\\
\textbf{"Pathological-Examination":} "Pathological examination 1: Routine: (left vocal cord tumor) sent for examination for mucosal squamous epithelial hyperplasia with excessive keratosis, and the local foci showed low-grade intraepithelial neoplasia. (Right vocal cord mass) sent for examination of high-grade intraepithelial neoplasia in mucosal squamous epithelium. Supplementary report 1: After clinical communication and review, the report is as follows: (left vocal cord tumor) was sent to the examination for mucosal squamous epithelial hyperplasia with excessive keratosis, and the local foci showed low-grade intraepithelial neoplasia. (Right vocal cord mass) sent for high-grade intraepithelial neoplasia in mucosal squamous epithelium (severe atypical dysplasia in squamous epithelium, partially carcinogenesis in situ).",\\
\textbf{"Diagnosis":} "Sinusitis, liver cysts, laryngeal cancer, tracheostomy state, prostate calcification foci, fatty liver, left kidney stones",\\
\textbf{"Options":} "A. Left kidney stones, B. Varicocele in the gastric fundus, C. Prostate calcification foci, D. Compound ulcer (scar stage), E. Tracheostomy state, F. Hepatic cyst, G. Sinusitis, H. Anemia, I. Laryngeal cancer, J. Fatty liver, K. Pneumothorax (a little)",\\
\textbf{"Label":} "A, C, E, F, G, I, J"\\
\end{longtable}
\end{onecolumn}
\clearpage

\begin{table*}[t]
\centering
\caption{Prompt for Outpatient Doctor.}
\label{tab:prompt1}
\begin{tabular}{p{16cm}}
\toprule
\textbf{Prompt for Outpatient Doctor} \\
\midrule
You are an experienced attending physician responsible for detailed inquiry of the patient's chief complaints, medical history, and initial physical examination. \\
Based on the retrieved medical knowledge: \\
\{retrieved\_info\} \\
\\

Patient information is as follows: \\
\{ \\
"Age": \{age\}, \\
"Sex": \{sex\}, \\
"Chief-Complaints": \{chief\_complaints\}, \\
"Present-Illness": \{present\_illness\}, \\
"Physical-Examination": \{physical\_examination\} \\
\} \\
\\

Please analyze the patient's information and output evidence tree structure in following format:\\
1. Clinical Clues: Identify key clinical clues present in the patient's chief complaints, medical history, and initial physical examination.\\
2. Possible Diseases: List possible diseases that these clinical clues might point to.\\
3. Reasoning Process: For each possible disease, provide a brief explanation of your reasoning process.\\
4. Evidence: For each possible disease, summarize the supporting evidence from the chief complaints, medical history, and physical examination.\\
\\
Output Format: \\

\dirtree{%
.1 Chief Complaints Clinical Reasoning Pathway.
.2 Disease 1.
.3 Analysis: Brief explanation of the reasoning process.
.4 Evidence 1: Chief Complaints.
.4 Evidence 2: Medical History.
.4 Evidence 3: Physical Examination.
.2 Disease 2.
.3 Analysis: Brief explanation of the reasoning process.
.4 Evidence 1: Chief Complaints.
.4 Evidence 2: Medical History.
.4 Evidence 3: Physical Examination.
.2 ...
}
Please ensure that the output strictly follows the above format and only includes the evidence tree structure. Avoid any additional text or explanations outside the tree structure.\\
\bottomrule
\end{tabular}
\end{table*}

\clearpage

\begin{table*}[t]
\centering
\caption{Prompt for Laboratory Doctor.}
\label{tab:prompt2}
\begin{tabular}{p{16cm}}
\toprule
\textbf{Prompt for Laboratory Doctor} \\
\midrule
You are an experienced laboratory physician responsible for analyzing laboratory test results and providing diagnostic suggestions based on the results.\\

Based on the retrieved medical knowledge:\\
\{retrieved\_info\}\\

The laboratory test results are as follows:\\
\{lab\_results\}\\

Please analyze the laboratory test results and output evidence tree structure in following format:\\
1. Abnormal Indicators: Identify which indicators in the laboratory test results are abnormal.\\
2. Possible Diseases: List possible diseases that these abnormal indicators might point to.\\
3. Reasoning Process: For each possible disease, provide a brief explanation of your reasoning process.\\
4. Evidence: For each possible disease, summarize the supporting evidence from the laboratory test results.\\
\\
Output Format: \\
\dirtree{%
.1 Laboratory Test Clinical Reasoning Pathway.
.2 Disease 1.
.3 Analysis: Brief explanation of the reasoning process.
.4 Evidence 1: Abnormal Indicator 1.
.4 Evidence 2: Abnormal Indicator 2.
.4 Evidence 3: Abnormal Indicator 3.
.2 Disease 2.
.3 Analysis: Brief explanation of the reasoning process.
.4 Evidence 1: Abnormal Indicator 1.
.4 Evidence 2: Abnormal Indicator 2.
.4 Evidence 3: Abnormal Indicator 3.
.2 ...
}

Please ensure that the output strictly follows the above format and only includes the evidence tree structure. Avoid any additional text or explanations outside the tree structure. \\

\bottomrule
\end{tabular}
\end{table*}

\begin{table*}[t]
\centering
\caption{Prompt for Radiology Doctor.}
\label{tab:prompt3}
\begin{tabular}{p{16cm}}
\toprule
\textbf{Prompt for Radiology Doctor} \\
\midrule
You are an experienced imaging physician responsible for analyzing imaging test results and providing diagnostic suggestions based on the results.\\

Based on the retrieved medical knowledge:\\
\{retrieved\_info\}

The imaging test results are as follows:\\
\{imaging\_results\}\\

Please analyze the imaging test results and output evidence tree structure in following format:\\
1. Abnormal Findings: Identify what abnormal findings are present in the imaging test results.\\
2. Possible Diseases: List possible diseases that these abnormal findings might point to.\\
3. Reasoning Process: For each possible disease, provide a brief explanation of your reasoning process.\\
4. Evidence: For each possible disease, summarize the supporting evidence from the imaging test results.\\
\\
Output Format: \\
\dirtree{%
.1 Imaging Test Clinical Reasoning Pathway.
.2 Disease 1.
.3 Analysis: Brief explanation of the reasoning process.
.4 Evidence 1: Abnormal Finding 1.
.4 Evidence 2: Abnormal Finding 2.
.4 Evidence 3: Abnormal Finding 3.
.2 Disease 2.
.3 Analysis: Brief explanation of the reasoning process.
.4 Evidence 1: Abnormal Finding 1.
.4 Evidence 2: Abnormal Finding 2.
.4 Evidence 3: Abnormal Finding 3.
.2 ...
}
Please ensure that the output strictly follows the above format and only includes the evidence tree structure. Avoid any additional text or explanations outside the tree structure.   \\
\bottomrule
\end{tabular}
\end{table*}

\begin{table*}[t]
\centering
\caption{Prompt for Pathology Doctor.}
\label{tab:prompt_pathology}
\begin{tabular}{p{16cm}}
\toprule
\textbf{Prompt for Pathology Doctor} \\
\midrule
You are an experienced pathology physician responsible for analyzing pathology test results and providing diagnostic suggestions based on the results.\\

Based on the retrieved medical knowledge:\\
\{retrieved\_info\}

The pathology test results are as follows:\\
\{pathology\_results\}\\

Please analyze the pathology test results and output evidence tree structure in following format:\\
1. Abnormal Findings: Identify what abnormal findings are present in the pathology test results.\\
2. Possible Diseases: List possible diseases that these abnormal findings might point to.\\
3. Reasoning Process: For each possible disease, provide a brief explanation of your reasoning process.\\
4. Evidence: For each possible disease, summarize the supporting evidence from the pathology test results.\\
\\
Output Format: \\
\dirtree{%
.1 Pathology Test Clinical Reasoning Pathway.
.2 Disease 1.
.3 Analysis: Brief explanation of the reasoning process.
.4 Evidence 1: Abnormal Finding 1.
.4 Evidence 2: Abnormal Finding 2.
.4 Evidence 3: Abnormal Finding 3.
.2 Disease 2.
.3 Analysis: Brief explanation of the reasoning process.
.4 Evidence 1: Abnormal Finding 1.
.4 Evidence 2: Abnormal Finding 2.
.4 Evidence 3: Abnormal Finding 3.
.2 ...
}
Please ensure that the output strictly follows the above format and only includes the evidence tree structure. Avoid any additional text or explanations outside the tree structure.   \\
\bottomrule
\end{tabular}
\vspace{10pt}
\end{table*}

\begin{table*}[t]
\centering
\caption{Prompt for Agent to Decide Whether to Participate in the Discussion.}
\label{tab:current_situation}
\begin{tabular}{p{16cm}}
\toprule
\textbf{Prompt for Agent to Decide Whether to Participate in the Discussion} \\
\midrule
Based on the current discussion situation, actively identify areas where your perspective differs from others'. Consider if providing your unique viewpoint could help resolve disagreements or improve the diagnosis. You should participate whenever there's an opportunity to clarify your position or persuade others, even if some opinions have already been expressed. \\
Do you need to provide new insights or engage in discussion with other doctors? \\
Please answer only with "Yes" or "No". \\

Current situation: \\
1. Patient case: \\
\{patient\_case\} \\
2. Current round: Round \{round\_num\}, Turn \{turn\_num\} \\
3. Diagnosis opinions from each doctor: \\
\{doctor\}: \{opinion\}
\\
\bottomrule
\end{tabular} 
\end{table*}

\begin{table*}[t]
\centering
\caption{Prompt for Generating Opinions for Specific Doctors.}
\label{tab:prompt_opinion}
\begin{tabular}{p{16cm}}
\toprule
\textbf{Prompt for Generating Opinions for Specific Doctors} \\
\midrule
As the \{source\_doctor\} doctor, please provide your professional opinion on the diagnosis from the \{target\_doctor\} doctor: \\
1. Patient case: \\
\{patient\_case\} \\
2. Current round: Round \{round\_num\}, Turn \{turn\_num\} \\
3. Diagnosis opinions from each doctor: \\
\{doctor\}: \{opinion\} \\
\\
Please concisely express your views, focusing on: \\
1. Which aspects of the other doctor's opinion you agree or disagree with \\
2. What additional insights or suggestions you have based on your expertise \\
3. How to integrate both professional perspectives to improve the diagnosis \\
\bottomrule
\end{tabular}
\vspace{10pt}
\end{table*}

\begin{table*}[t]
\centering
\caption{Prompt for Collecting and Updating Viewpoints.}
\label{tab:prompt_updated_tree}
\begin{tabular}{p{16cm}}
\toprule
\textbf{Prompt for Collecting and Updating Viewpoints} \\
\midrule
As the \{doctor\_type\} doctor, please generate an UPDATED diagnostic tree based on original assessment and new feedback. \\
\\
Output Format: \\
\dirtree{%
.1 \{doctor\_type\} Doctor Reasoning Pathway.
.2 Disease 1.
.3 Analysis: Brief explanation of the reasoning process.
.4 Evidence 1: Abnormal Finding 1.
.4 Evidence 2: Abnormal Finding 2.
.4 Evidence 3: Abnormal Finding 3.
.2 Disease 2.
.3 Analysis: Brief explanation of the reasoning process.
.4 Evidence 1: Abnormal Finding 1.
.4 Evidence 2: Abnormal Finding 2.
.4 Evidence 3: Abnormal Finding 3.
.2 ...
}
Please ensure that the output strictly follows the above format and only includes the evidence tree structure. Avoid any additional text or explanations outside the tree structure. \\
\\
1. Original diagnosis: \\
\{self.round\_opinions[1][doctor\_type]\} \\
2. Feedback received in this round: \\

\bottomrule
\end{tabular}
\end{table*}

\begin{table*}[t]
\centering
\caption{Prompt for Summarizing and Obtaining Final Opinions.}
\label{tab:prompt_final_diagnosis}
\begin{tabular}{p{16cm}}
\toprule
\textbf{Prompt for Summarizing and Obtaining Final Opinions} \\
\midrule
As the head of the medical team, please make the final diagnosis based on the following information: \\
\\
1. Patient case: \\
\{json.dumps(self.patient\_case[0], ensure\_ascii=False, indent=2)\} \\
2. Diagnosis opinions from the last round: \\
\{doctor\}: \{opinion\} \\
3. Diagnosis options: \\
\{self.options\} \\
\\
The output should include: \\
1. The final diagnosis result (please select the appropriate letter from the options) \\
2. The evidence tree structure, formatted as follows: \\
\dirtree{%
.1 Reasoning Pathway.
.2 Disease 1.
.3 Analysis: Brief explanation of the reasoning process.
.4 Evidence 1: Abnormal Finding 1.
.4 Evidence 2: Abnormal Finding 2.
.4 Evidence 3: Abnormal Finding 3.
.2 Disease 2.
.3 Analysis: Brief explanation of the reasoning process.
.4 Evidence 1: Abnormal Finding 1.
.4 Evidence 2: Abnormal Finding 2.
.4 Evidence 3: Abnormal Finding 3.
.2 ...
}
The result should be output in JSON format, strictly following the format below. Do not add any extraneous words! \\
\begin{verbatim}
{
    "selected_options":"",
    "evi_tree":""
}
\end{verbatim}\\
\bottomrule
\end{tabular}
\end{table*}

\end{document}